\documentclass[a4paper, 10pt, twocolumn, twoside]{article}

\usepackage{assetowner}

\usepackage{lscape}
\usepackage{hologo}

\begin{document}
 % Do not change the following line
\linespread{0.5}

\title{Asset Ownership Identification: Using machine learning to predict enterprise asset ownership}

\author{Craig Jacobik}
%\email{\href{mailto:cjacobik3@gatech.edu}{cjacobik3@gatech.edu}}

% Do not change the following three lines
\maketitle 
\thispagestyle{fancy} 
\pagestyle{fancy}

\begin{abstract}
Asset owner identification is an important first step for any information security organization, allowing organizations the ability to identify and detect data breaches and losses, vulnerabilities, possible attack surfaces, and define effective countermeasures. Using existing asset ownership data, the research utilized an assortment of machine learning algorithms to determine the best classification model to predict an asset’s owner.    The research ran separate analyses for each enumerated team, then ran a 100 iteration Monte Carlo Cross Validation across Adaboost, Logistic Regression, Naïve Bayes, Classification and Regression Trees, and Random Forests. Finally, a visualization dashboard was created to help users understand the asset inventory through interactive exploratory data analysis as well as the ability to understand model evaluation metrics including accuracy, sensitivity, and specificity for each model. Overall, Adaboost performed the best across all owners with low testing errors below 5\% while Naïve Bayes performed the worst. The remaining models performed similarly. The fully qualified domain name (FQDN), Classless Inter-Domain Routing (CIDR) CIDR/16, and location were among the most important features.
\end{abstract}


\section{Motivation}
\label{sec:Motivation}
Enterprise asset ownership and asset identification is one of the first steps for any Information Security program.  The International Organization for Standardization (ISO) and International Electrotechnical Commission (IEC) jointly publish the ISO 27000 series of standards, which are some of the preeminent best practices for Information Security. One of the core standards details that asset identification and asset ownership are imperative to identify data breaches and losses, vulnerabilities, possible attack surfaces, and defining effective countermeasures \cite{beckers2011}. To ensure best practices, a recommended 7-step approach for compliance includes: refine assets including understanding the location and team responsible, understand threats and vulnerabilities, assess risks including business impact and security risk severity, instantiate security requirements, build controls, and generate documentation \cite{alebrahim2014}. 

 Asset ownership is a difficult problem. Luckily, there are a few tools that can be deployed to assist with the problem. There are Desktop Asset Management (DAM) tools available to assist in the inventory management, asset monitoring, and software distribution to these assets \cite{chou2002}. Additionally, Configuration Management Databases (CMDB) allow organizations the ability to discover, store, and track assets. Keller et al. discusses this best by identifying the optimal operational architecture and processes to ensure success when deploying a CMDB with relevant use cases from client engagements \cite{keller2009}. 

This research partnered with a large multinational datacenter company. Despite the use of the aforementioned tools, asset ownership remains a key concern and issue for the datacenter company as well as any organization. First, there are legacy naming conventions as well as newly adopted standards on asset naming conventions. While teams are encouraged and mandated to follow the updated standards, deviations do occur. Deviations occur across teams, geographies, datacenters, and more. Second, teams can quickly deploy virtual machines and cloud infrastructure with the touch of a button. With the intuitive and ease of deploying to the cloud, organizations, teams, and users can quickly deploy computing infrastructure. Third, large Internet Protocol (IP) subnets complicate the asset ownership problem. Sometimes teams will deploy assets into a larger region subnet such as the Data Science team deploying a server into the Americas’ IP subnet despite being in India. Fourth, as the company  has grown, it has acquired other datacenter companies. Acquisitions are a fantastic avenue of growth, but with acquisitions, come established servers, team ownership, asset naming conventions, and more IP subnets. Fifth, leaders and employees come and go. As such, owners or teams who may have had certain assets may no longer be with the company. Sixth and finally, while tools similar to DAM tools are available and deployed, numerous inconsistently deployed tools can create gaps within the asset inventory and ownership. The siloed approach can lead organizations to believe all assets are included in the DAM or CMDB while they may not exist.

Using existing asset ownership data exported for a sample of the asset inventory, the research utilized an assortment of machine learning algorithms to determine the model that created the best classification model to predict an asset’s owner. This approach attempted to fit models and predictions for each identified asset owner. Finally, a visualization dashboard was created to allow the client the opportunity to understand their asset inventory through intuitive Exploratory Data Analysis (EDA) and the ability to understand the accuracy, sensitivity, and specificity metrics of each model.

\section{Data}
\label{sec:Data}

The data consisted of a sampled export of a CMDB with nearly seventy thousand rows across nineteen columns. There was a fair amount of data cleaning that needed to be completed. The tags column consisted of a list of tags that would be used as features for the analysis. Therefore, it was necessary to parse and extract the relevant tags and assign those tags to the appropriate features. Combining the parsed tags as well as the other attributes, the most relevant features consisted of technical details of assets including the asset name, Netbios, operating system (OS), class name, fully qualified domain name (FQDN), IP address, Media Access Control (MAC) address, whether a CMDB agent is deployed, the regional location of the asset, the type of OS system, the owner of the system, and tags associated with the asset.   

For the initial analysis, a few additional fields were added as part of feature engineering to assist with the analysis. First, the FQDN was split to only include the most top-level information. For example, a hostname of ab1.cd2.corp.company.com was shortened to remove the datacenter and project information. The resulting FQDN would be corp.company.com in this example. Second, the IP address derived additional Classless Inter-Domain Routing (CIDR) features including CIDR/8, CIDR/16, and CIDR/24. For example, the CIDR/8 represents the first eight bits of the 32 bit IP address; or more concisely, the first octet of the four octets. Finally, the Organizational Unique Identifier (OUI), administered and assigned by the Institute of Electrical and Electronic Engineers (IEEE), was extracted from the MAC addresses. The OUIs were extracted using Wireshark’s Github repository \cite{github} and represent the first three bytes of the six byte MAC address.

\begin{figure*}[!htb]
    \centering
    \includegraphics[width=\textwidth]{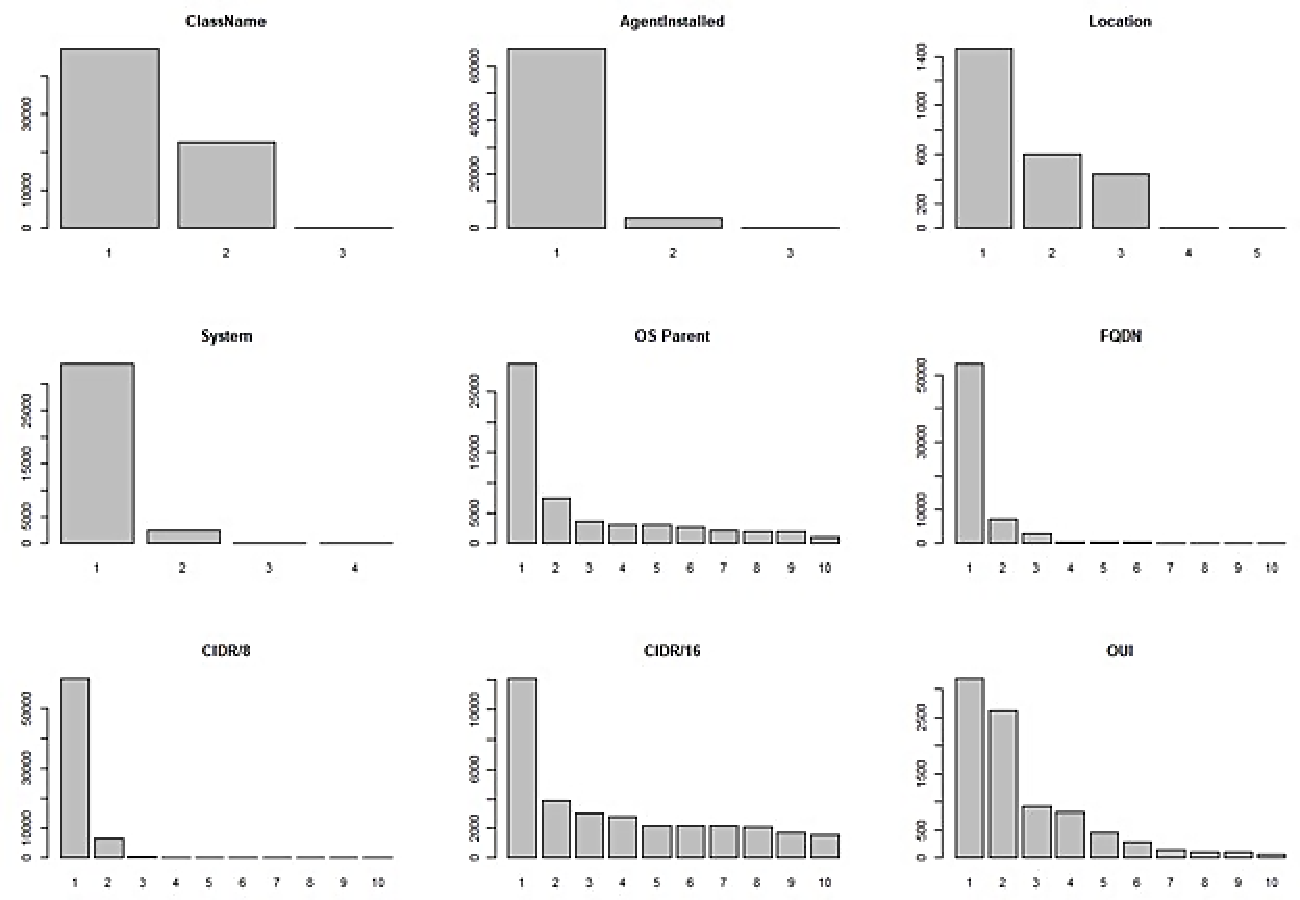}
    \caption{Frequency Barcharts showing distribution of variables for selected features}
    \label{fig_1}
\end{figure*}

The data variables were categorical in nature, which led to the selection of models to predict the classification response variables. Frequency barcharts are represented in Figure~\ref{fig_1} for a few of the features. Unsurprisingly, Linux assets were the most prevalent within the OS Parent feature. Within most features e.g. Agent Installed, System, FQDN, CIDR/8; there are 1-2 variables that have large concentrations with the remaining in the feature set having mostly uniform distributions. The rest of the features e.g. ClassName, Location, CIDR/16, OUI; had somewhat uniform distributions across their sets.

The code iterated through all of the enumerated teams to produce the response variable. For example, if team A was listed as the owner in the dataset, the response variable would have represented whether the asset was owned by team A. This meant that the models and Monte Carlo Cross Validation techniques detailed in the Methodology section below were run dozens of times. Further, this meant there was an unbalanced response variables for certain team predictions.

\section{Methodology}
\label{sec:Methodology}
\subsection{Modeling}
The problem was modeled as a binary classification. For each owner, the response variable was either that owner or not.     To test for robustness and minimize sampling bias, the analysis ran every owner response through 100 iterations of a Monte Carlo Cross Validation approach. Each iteration produced a separate training set (80\%) to train the model, a cross-validation set (10\%) for testing hyperparameter selection where appropriate, and a hold-out test set (10\%) for the final selected model to be evaluated against. In general, binary classification problems are solved by a variety of methods but the following were selected:

\subsubsection{Adaboost or Gradient Boosted Machines}
Adaboost is an iterative reweighting boosting tree-based approach that starts with weak learners and then iterates through, reweighting after each round focusing on previously misclassified points \cite{freund1997}\cite{schapire2013}. There is limited research concerning the applicability of Adaboost to the Asset Owner problem, but there exists numerous examples of its applicability in similar types of problems including those cited by Hatwell et al. \cite{hatwell2020}.

The analysis trained the model to the dataset tuning the hyperparameters including shrinkage which represents the learning rate for each tree in the expansion and the depth parameter which represent the max depth of each tree. The parameters included shrinkage as seen in Equation~\ref{eq_1} and depth as seen in Equation~\ref{eq_2}. The parameters were evaluated against the cross-validation set and the optimal hyperparameters were then evaluated against the holdout test set.
\begin{equation}\label{eq_1}
shrinkage \in \{0.01,0.05,0.1\}
\end{equation}
\begin{equation}\label{eq_2}
depth \in \{2,4,6,8\}
\end{equation}

\subsubsection{Logistic Regression}
Logistic regression is a classical and frequently used statistical implementation of a logit function transformation to map binary response variables to the regression model. When utilizing logistic regression, it is necessary to validate several assumptions: logit transformation is a linear combination of predicting variables, independent response variables, there should be little or no multicollinearity between predictor variables, and finally that the link function is a s-shaped logit function. Assuming those assumptions hold, the logistic regression is a valid model for the data.

Logistic Regression has been extensively studied and applied to many different problems. Komarek talks about numerous applications of logistic regression \cite{komarek2004}. Specific to Information Security, successful applications and papers exist around the effective use of logistic regression for Internet of Things (IoT) and Fog computing asset management \cite{teoh2021}. IoT is a small subset of the scope of the asset owner identification problem. 
\subsubsection{Naïve Bayes}
Naïve Bayes is a probabilistic decision model that utilizes Bayes Rule and the assumption of independent variables \cite{murphy2006}. For categorical variables, similar to the asset owner identification problem, required probabilities use frequency counts and probability estimates derived from smoothing functions such as Laplace estimate \cite{webb2010}. The analysis tuned the Laplace smoothing parameter as seen in Equation~\ref{eq_3}, selecting the optimal value based on the results on the cross-validation set. 
\begin{equation}\label{eq_3}
laplace \in \{0,1\}
\end{equation}

Naïve Bayes has been applied against myriads of information security types of problems. It has been applied for risk assessment \cite{foroughi2008} as well as classifying network enabled devices \cite{arora2016}. Specific to the latter application, asset identification is necessary to allow organizations to remediate security incidents quickly and accurately \cite{arora2016}. The paper by Arora et al. evaluated 4 approaches including K-Nearest-Neighbor (KNN), Naïve Bayes, Support Vector Machines (SVM), and Random Forests (RF) algorithms with Random Forests performing best and Naïve Bayes performing the worst \cite{arora2016}. Given that the asset owner identification research presented used two of these four models (Naïve Bayes and Random Forests), it could be expected that the results mirror the results by Arora et al.

\subsubsection{Classification and Regression Trees (CART)}
CART leverages decision trees which split based on the data impurity measures (gini index or entropy) to classify the response \cite{breiman1984}. With over fifty years of application, CART models remain one of the most popular and applied techniques \cite{loh2014}. Within CART, the analysis tuned and evaluated the complexity parameter for minimum improvement as seen in Equation~\ref{eq_4}, minimum number of observations needed in the node to split as seen in Equation~\ref{eq_5}, and the maximum depth of any node in the tree as seen in Equation~\ref{eq_6}. CART models are great nonlinear models that should perform similarly to Adaboost and Random Forests algorithms for the asset owner identification problem.
\begin{equation}\label{eq_4}
cp \in \{0.01,0.05,0.1\}
\end{equation}
\begin{equation}\label{eq_5}
minsplit \in \{5,10,15,20\}
\end{equation}
\begin{equation}\label{eq_6}
maxdepth \in \{2,5,10,20\}
\end{equation}

\subsubsection{Random Forests}
Random Forests is an ensemble method that uses an iterative random selection of features to split each node and create many tree models \cite{breiman2001}. Random Forests models compare favorably with other bagging and boosting models such as Adaboost \cite{breiman2001}. Internal metrics measure error, strength, and correlation allowing the Random Forests to measure variable importance \cite{breiman2001}. Within Random Forests, there are a few hyperparameters to tune including the number of trees as seen in Equation~\ref{eq_7}, the number of variables randomly sampled at each split as seen in Equation~\ref{eq_8}, and the maximum number of terminal nodes as seen in Equation~\ref{eq_9}.
\begin{equation}\label{eq_7}
ntree \in \{250,500,1000,2000\}
\end{equation}
\begin{equation}\label{eq_8}
mtry \in \{2,3,4,5\}
\end{equation}
\begin{equation}\label{eq_9}
maxnodes \in \{3,5,8,10,15\}
\end{equation}

Random Forests are used extensively within the Information Security domain. Random Forests was one of the algorithms evaluated for Asset Identification \cite{arora2016}. It was also used to create a multi-dimensional network asset identification model \cite{xiong2021}. Lastly, Random Forests and the Isolation Forest algorithm was used to manage and detect rogue assets thereby strengthening security postures \cite{adebayo2022}. Given the prevalent use of Random Forests for asset identification, it is an appropriate model to use for asset ownership identification.

\subsection{Visualization}
The final deliverable should include an interactive visualization dashboard. Ideally, it would be broken into two parts. The first part would allow users the ability to perform high-level and real-time exploratory data analysis. For each of the features, the dashboard should have bubble charts, barcharts, and colored tables showing the frequencies of the important variables. Additionally, users should have the ability to filter variables, thereby updating the graphs. For example, if a user selects the Americas regions, all the remaining visualizations should update their values and graphics to represent what occurs in the Americas. Advanced users should also be able to filter using complex boolean logic, allowing users the ability to understand very specific details about the assets.

The second part of the dashboard should include information about the modeling results. Users must select an owner response variable that they are interested in e.g. Data Science team. From there, users should be able to see an aggregated confusion matrix to understand the difference between each of the five model’s predictions and the actual values. This would allow users to understand an array of performance metrics including accuracy, specification, specificity, F1 score, and more. Additionally, there should be a graphical depiction of the performance metrics across the models. For example, if a user wants to see the accuracy percentage across all five models, a barchart may be a good representation of how each of the models performed. Finally, a user should be able to see the data in the test set as well as each model’s owner classification and the actual classification. This will allow users the ability to impute what a theoretical asset may produce. The user could run the data through the model afterwards. Most importantly, these visualizations should be filterable such that if a user wanted to see how the Random Forests model did at predicting, all the visualizations would update accordingly.

\section{Results}
\label{sec:Results}
\subsection{Modeling}
For a selection of asset owners, the model results from the test set are seen below in Figure~\ref{fig_2}. One thing is clear; Adaboost (Model 1) consistently performed the best by having the lowest median testing error rates across all Monte Carlo Cross Validation runs. Additionally, Adaboost appears to have the smallest, or near the smallest, interquartile range signifying the overall robustness of the model choice. The error rates for Adaboost were consistently less than 5\% implying an accuracy rate of 95\% or better. Naïve Bayes appears to perform consistently the worst. These results correspond to the outcomes from Arora et al. \cite{arora2016}.

\begin{figure*}[!htb]
    \centering
    \includegraphics[width=\textwidth]{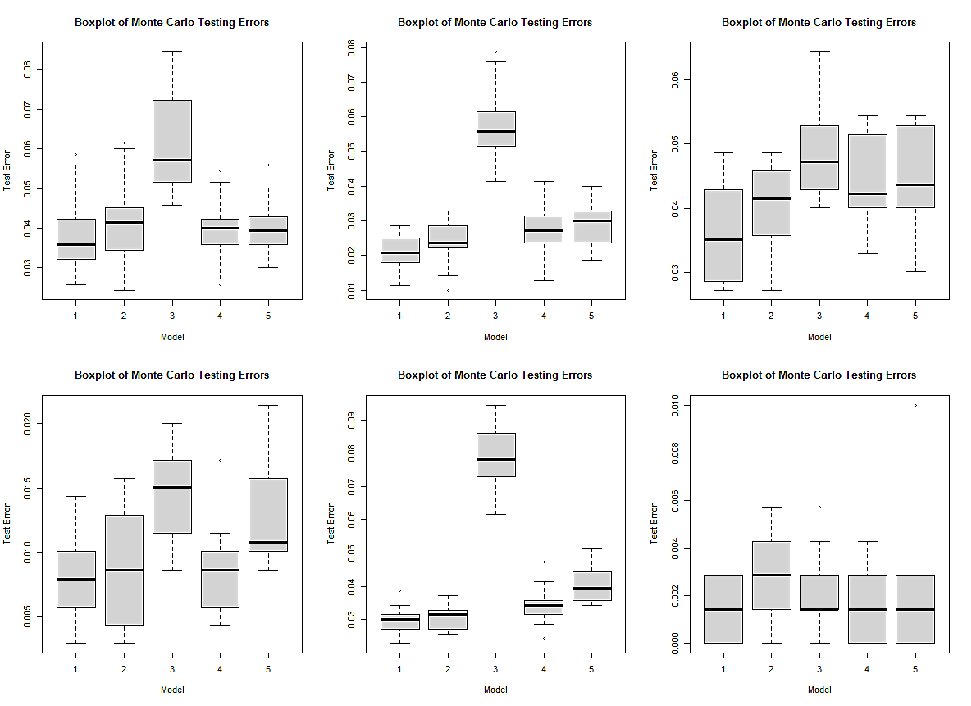}
    \caption{Monte Carlo Cross Validation Error Rates for five models across selection of response variables}
    \label{fig_2}
\end{figure*}

Looking at the features across all the models, it appears that FQDN, CIDR/16, Location, OUI, and operating system parent are consistently the most important features. This is true across the Adaboost, CART, and Random Forests models.
\subsection{Visualization}
As mentioned in the Visualization subsection of the Methodology section, the sample dashboard allows for exploratory data analysis and it also allows for users to delve deeper into the model results. Certain data has been intentionally omitted from Figure~\ref{fig_3} below.

The final deliverable included an interactive visualization dashboard. This dashboard allows users the ability to perform high-level and real-time exploratory data analysis. Bubble charts show the distribution of assets by team. Also, there are bubble charts showing the distribution of the CIDR/16 assets colored and grouped by the CIDR/8 subnets. The final bubble charts show the operating systems colored and grouped by the parent e.g. Linux kernel 2.4 and Linux kernel 3.2 would be grouped under the linux parent. Additionally, the visualization has colored tables showing the frequency of values for each of the following variables: ClassName, Agent Installed, Location, FQDN, OS Parent, OUI, and CIDR/8. All variables can be filtered with the graphs and metrics updating appropriately.

The second half of the dashboard includes information about the modeling results. Users must select an owner response variable that they are interested in. From there, an extended confusion matrix is available to understand the difference between the model predictions and the actual values. It is possible to acquire performance metrics including accuracy, specification, specificity, F1 score, and more. A barchart also shows the accuracy of each of the five models. Additionally, there is a table with the test dataset available including the features used in the prediction and the results of the five algorithms and the actual response variable. The models as well as the response variable are filterable i.e. if a user wanted to see how all models did against the true response variable, they could ascertain that information.

\begin{figure*}[!htb]
    \centering
    \includegraphics[width=\textwidth]{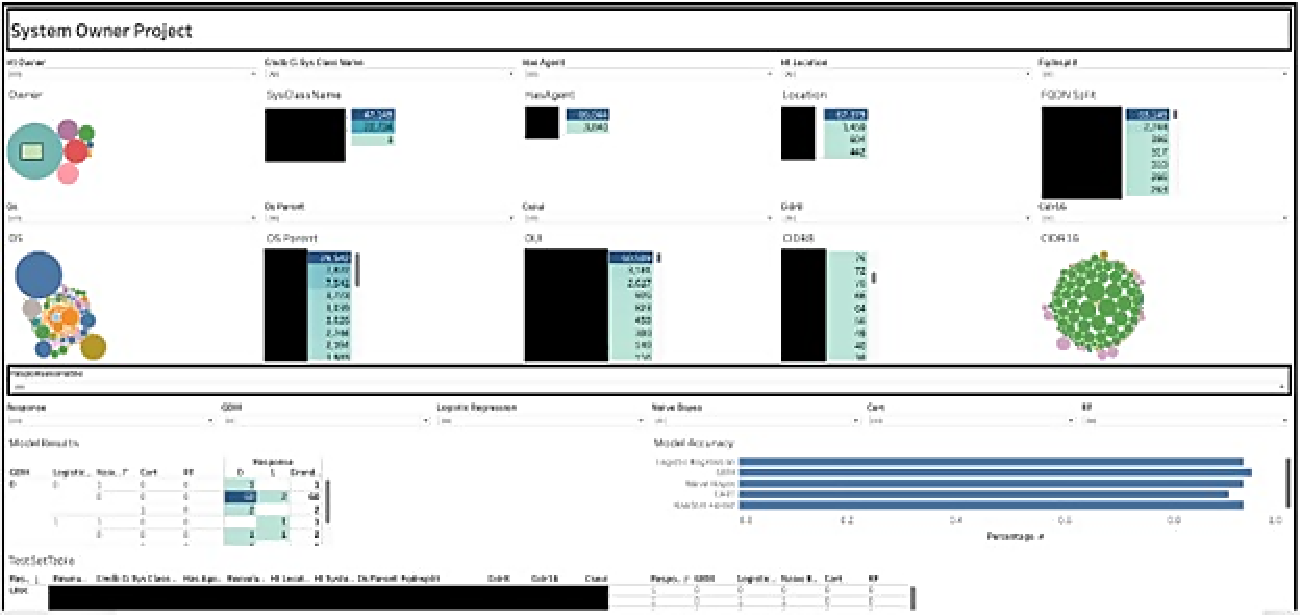}
    \caption{Visualization dashboard}
    \label{fig_3}
\end{figure*}

\section{Conclusion}
\label{sec:Conclusion}
Asset owner identification is an important first step for any information security organization  as doing so allows organizations the ability to identify and detect data breaches and losses, vulnerabilities, possible attack surfaces, and define effective countermeasures. Using existing asset ownership data, the research utilized an assortment of machine learning algorithms to determine the best classification model to predict an asset’s owner. The research ran separate analyses for each owner. Within each analysis, a 100 iteration Monte Carlo Cross Validation was run across a litany of modeling algorithms including  Adaboost, Logistic Regression, Naïve Bayes, Decision Trees, and Random Forests. Finally, a visualization dashboard was created that allows users the ability to perform exploratory data analysis and consume the model outputs and metrics.

Overall, Adaboost performed the best with low testing errors below 5\%. Generally, ensemble methods for classification produce more accurate results than classical approaches. While the model is more difficult to interpret, the iterative reweighting tree-based approach that focuses on misclassified data clearly performs the best. Naïve Bayes performed the worst of the five models. Some of the assumptions such as independence between predictors may have led to its demise. The remaining three models of Logistic Regression, CART, and Random Forests all performed well but not as well as Adaboost. Finally, FQDN, CIDR/16, Location were consistently the most important features within the models.

Future research could revolve around additional feature engineering, testing additional asset inventories, attempting other modeling algorithms, and enhancing the visualization dashboard. Nonetheless, the current research performed better than expected on the dataset. The intuitive and interactive dashboard allowed users a way to understand data insights previously unavailable to them and the predictive modeling allowed for previously unlabeled assets to be assigned a predicted asset owner . This research helped the datacenter company predict owners for unidentified assets. Additionally, the company used this research to compare and validate the current ownership for identified assets.

%\bibliography{assetowner}

\end{document}